\documentclass[10pt,twocolumn,letterpaper]{article}

\usepackage{iccv}
\usepackage{times}
\usepackage{epsfig}
\usepackage{graphicx}
\usepackage{amsmath}
\usepackage{amssymb}
\usepackage{enumitem}

\usepackage[pagebackref=true,breaklinks=true,letterpaper=true,colorlinks,bookmarks=false]{hyperref}

\iccvfinalcopy 

\def\iccvPaperID{10191} 

\ificcvfinal\fi

\begin{document}

\title{MetaCAM: Ensemble-Based Class Activation Map}

\author{Emily Kaczmarek\textsuperscript{a,*}, Olivier X. Miguel\textsuperscript{b}, Alexa C. Bowie\textsuperscript{b}, Robin Ducharme\textsuperscript{b}, Alysha L.J. Dingwall-Harvey\textsuperscript{a,b}, \\Steven Hawken\textsuperscript{a,b,c,g}, Christine M. Armour\textsuperscript{a,i,j}, Mark C. Walker\textsuperscript{a,b,c,d,e,f,g}, Kevin Dick\textsuperscript{a,*}\\
\textsuperscript{a} Children’s Hospital of Eastern Ontario Research Institute, Ottawa, Canada\\
\textsuperscript{b} Clinical Epidemiology Program, Ottawa Hospital Research Institute, Ottawa, Canada\\
\textsuperscript{c} School of Epidemiology and Public Health, University of Ottawa, Ottawa, Canada\\
\textsuperscript{d} Department of Obstetrics and Gynecology, University of Ottawa, Ottawa, Canada\\
\textsuperscript{e} International and Global Health Office, University of Ottawa, Ottawa, Canada\\
\textsuperscript{f} BORN Ontario, Children’s Hospital of Eastern Ontario, Ottawa, Canada\\
\textsuperscript{g} ICES, Toronto, Canada\\
\textsuperscript{h} Department of Obstetrics, Gynecology \& Newborn Care, The Ottawa Hospital, Ottawa, Canada\\
\textsuperscript{i} Department of Pediatrics, University of Ottawa, Ottawa, Canada\\
Institution1 address\\
\textsuperscript{j} Prenatal Screening Ontario, Better Outcomes Registry \& Network, Ottawa, Canada\\
{\tt\small \textsuperscript{*}\{ekaczmarek, kdick\}@cheo.on.ca}
}

\maketitle
\ificcvfinal\thispagestyle{empty}\fi

\begin{abstract}\vspace{-3mm}
The need for clear, trustworthy explanations of deep learning model predictions is essential for high-criticality fields, such as medicine and biometric identification. Class Activation Maps (CAMs) are an increasingly popular category of visual explanation methods for Convolutional Neural Networks (CNNs).  However, the performance of individual CAMs depends largely on experimental parameters such as the selected image, target class, and model. Here, we propose MetaCAM, an ensemble-based method for combining multiple existing CAM methods based on the consensus of the top-$k$\% most highly activated pixels across component CAMs. We perform experiments to quantifiably determine the optimal combination of 11 CAMs for a given MetaCAM experiment. A new method denoted Cumulative Residual Effect (CRE) is proposed to summarize large-scale ensemble-based experiments. We also present adaptive thresholding and demonstrate how it can be applied to individual CAMs to improve their performance, measured using pixel perturbation method Remove and Debias (ROAD). Lastly, we show that MetaCAM outperforms existing CAMs and refines the most salient regions of images used for model predictions. In a specific example, MetaCAM improved ROAD performance to 0.393 compared to 11 individual CAMs with ranges from -0.101-0.172, demonstrating the importance of combining CAMs through an ensembling method and adaptive thresholding.  


\end{abstract}

\section{Introduction}

Convolutional neural networks (CNNs) are state-of-the-art deep learning architectures developed for image analysis tasks, including classification, segmentation, and object detection. These methods were originally considered uninterpretable (\textit{i.e.} `black boxes') given that specific regions or features of an image used to produce a model’s prediction were unknown. Having clear, reliable model interpretation improves confidence and trust in deploying artificial intelligence in real-world settings. This is of particular importance in high-criticality fields, such as medicine, autonomous driving, and automatic biometric identification \cite{Schmid21}. Furthermore, interpretability can identify decisions using incorrect information, such as biases or undesired markings in images. There is a definitive need for dependable visualizations of salient regions used in predictions. 

Numerous studies have investigated improving the explainability of CNNs. An increasingly popular class of explainable algorithms for interpreting CNN model predictions are Class Activation Maps (CAMs) \cite{Zhou16}. Originally developed by Zhou \textit{et al.}, CAMs create heat-map visualizations indicating the most salient regions of an image used by a CNN model for a given task. These visualizations are typically generated through a linear weighting of the feature maps produced by the final convolutional layer of a network. Many variations have been propose to improve upon the original CAM formulation \cite{Chattopadhay18,Desai20,Draelos20,Fu20,Gildenblat21,Jiang21,Muhammad20,Selvaraju17,Wang20}. There is, however, little consensus regarding which CAM method produces the most accurate reflection of important regions within an image. 

The comparison across different CAM variants has also been largely inconsistent, both with what CAM methods are compared and what performance metric is considered. Qualitatively, performance may be evaluated in a given study by comparing visualizations between various CAMs. Quantitatively, various performance metrics have been proposed including perturbation analysis, object localization and segmentation, and human trust/class discrimination, making relative CAM ranking infeasible. Furthermore, the performance of CAM methods varies across the parameters of individual experiments, such as the chosen images, their target classes, and the CNN model. The combination of CAM visualizations for improved performance and consistency has recently been investigated \cite{Ornek23}; however, the selection of contributing CAMs is arbitrary and understudied.

In this work, we address these issues by proposing a generalized method MetaCAM, a consensus-based CAM method that outputs the top-$k$\% pixels in agreement across any number and combination of component CAM methods. This consensus-based approach ensures that if a particular CAM performs poorly for a specific task, its contribution will be mediated in the final MetaCAM visualization. Consequently, MetaCAM may be used reliably in a diverse application to generate valid CAM visualization depicting the most salient regions of an image.

We additionally develop an adaptive thresholding method to determine optimal top-$k$ values for maximizing MetaCAM performance. We further extend this method to refine and improve existing CAM methods. Through large-scale comparative experiments, we systematically determine what combinations of component CAMs should be considered as part of MetaCAM among 11 unique publicly implemented CAM methods
 \cite{Gildenblat21}.

\vspace{8pt}
\noindent Our key contributions are as follows:

\begin{itemize}
\item We propose MetaCAM, a novel ensemble-based CAM method that combines existing CAMs; we perform extensive experimentation to determine the best aggregation of CAM methods.

\item We demonstrate that MetaCAM can be extended to include non-CAM visual explanation methods such as FullGrad \cite{Srinivas19}. For simplicity, when referring to CAM methods throughout this paper, we additionally subsume FullGrad.

\item We develop adaptive thresholding to improve the performance of MetaCAM. We further demonstrate how this can be extended to individual CAM methods to greatly improve performance and refine visualizations. 

\item We perform a systematic evaluation of MetaCAM combinations of 11 individual CAM methods. We summarize performance across MetaCAM and all individual CAMs for numerous images, target classes, and CNN models using an unbiased quantitative performance metric, Remove and Debias (ROAD).
\end{itemize}

\section{Related Work}

\subsection{CAM Methods}

For a given CAM method, the CAM visualization \textit{L} is generated from a linearly weighted summation of all \textit{k} feature maps \textit{A} at the chosen layer \textit{l} of a CNN architecture \textit{f}. Each CAM method produces a map for a given image, \textit{x}, and class-discriminative methods further specify the desired class output, \textit{c}. Most CAM methods also perform a ReLU operation after the final summation to retain positive activations (not included in Eq. \ref{eqn:cam-form}). The CAM formulation is thus:
\vspace{-12px}
\begin{equation}\label{eqn:cam-form}
L^c_{CAM(A)} = \sum\limits_{k}(\alpha_k^cA_k),  \textit{where }  A= f ^l(x)
\end{equation}

The original CAM determined \(\alpha_k^c\), the importance of individual feature maps, using the weights of the final dense layer leading to class predictions in a CNN \cite{Zhou16}. However, this required the final dense layer to be preceded by a convolutional layer and global average pooling, which can reduce performance and is not included in many modern CNNs. As such, new CAM methods have been developed that are CNN model-agnostic and have improved upon existing CAM performance. We have grouped 11 existing CAM variants into distinct categories based on how they compute Eq. \ref{eqn:cam-form}. 

\textbf{Basic GradCAMs:} GradCAM was the first method to propose the use of gradients to determine the importance of feature maps \cite{Selvaraju17}. Specifically, the gradient of the desired class with respect to the feature maps at a specified layer is used as the \(\alpha_k^c\) weights in Eq. \ref{eqn:cam-form}. This eliminated the need for a specific network architecture and demonstrated improved performance over the original CAM computation. GradCAM++ is a variation of GradCAM which leverages a weighted average of first-order gradients based on higher-order gradients \cite{Chattopadhay18}. While these two methods are consistently used across studies and are highly popularised, certain critiques (\textit{e.g.} \cite{Draelos20}) have led to the development of further CAM variants. 

\textbf{GradCAM Variants:} Different GradCAM variants have been proposed to further improve the performance of GradCAM and GradCAM++ and address some of their limitations. Fu \textit{et al.} argue that there is little theoretical explanation to explain averaging gradients in GradCAM \cite{Fu20}. Instead, they propose XGradCAM, which calculates a weighted average of gradients determined through an optimization problem based on axiom constraints of sensitivity and conservation. EigenGradCAM is also a GradCAM variant that has also been implemented by Gildenblat as a class-discriminative variation of EigenCAM \cite{Gildenblat21,Muhammad20}. In this case, \(\alpha_k^c\) remains as the gradient, but rather than multiplying this by the feature map activations \textit{A}, the principal components of the activations are instead used. 

\begin{figure*}
    \centering
    \includegraphics[width=\textwidth]{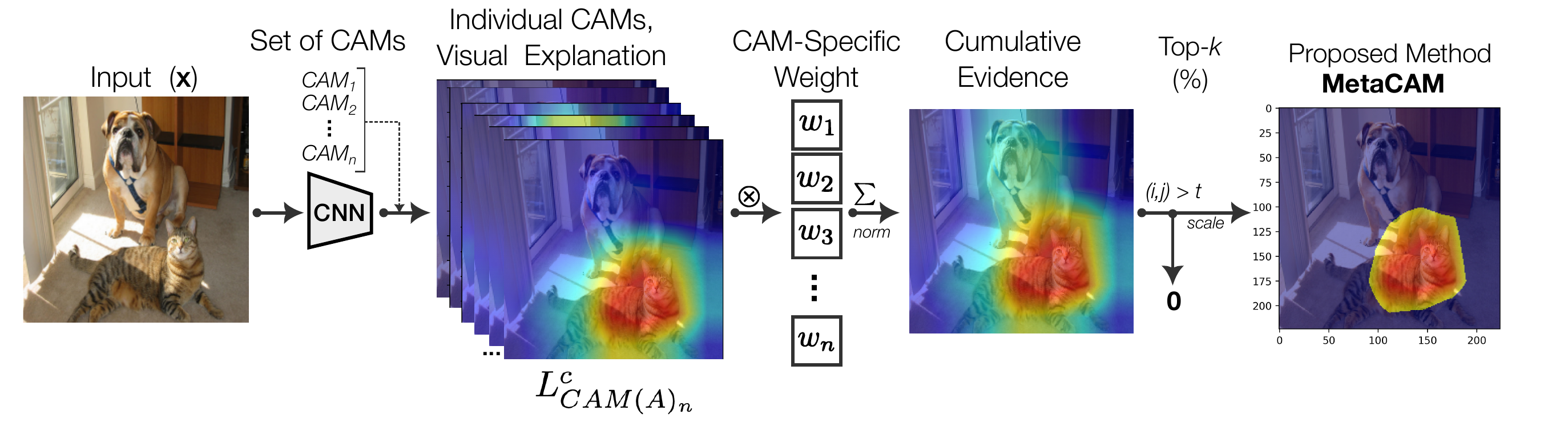}
    \caption{Conceptual overview of the proposed MetaCAM visual explanation method. For clarity, $w=1$ when thresholding is used ($k < 100$\%) and used otherwise when $k=100$\%.}
    \label{fig:conceptual-overview}
\end{figure*}

\textbf{Elementwise GradCAMs:} Rather than weighting feature maps by the average of first or second gradients, some methods suggest performing elementwise multiplication of gradients by feature maps. Draelos and Carin show that averaging gradients may cause certain image elements to falsely appear as regions of importance \cite{Draelos20}. For example, negative elementwise gradients may become positive from averaging and therefore incorrectly highlight areas on their respective feature maps. To avoid this, HiResCAM multiplies feature maps by their elementwise gradients, which ensures each pixel is weighted for importance by their respective gradient value \cite{Draelos20}. A similar elementwise CAM implementation, GradCAMElementwise, was proposed by Gildenblat \cite{Gildenblat21}. In this case, after performing elementwise multiplication of feature maps by gradients, ReLU is employed prior to the summation of feature maps.

\textbf{LayerCAM:} Jiang \textit{et al.} recognized that the final layers of a CNN produce coarse feature maps, which may cause the CAM visualizations to lose important fine-detailed information \cite{Jiang21}. To counter this, LayerCAM feature map activations are weighted using spatial-specific gradients (elementwise gradients), using positive gradients only. This ensures detailed information can be captured from any layer in the CNN, and these layers can be summed together for more specific CAM visualizations.

\textbf{Image Perturbation-Based Approaches:} While high performance has been achieved using gradient-based CAM methods, studies suggest that gradient saturation may lead to noisy or diminished visualizations. Wang \textit{et al.} also show that gradient-based methods may result in incorrectly weighted feature maps \cite{Wang20}. ScoreCAM and AblationCAM are two alternative approaches that use perturbations to identify feature map importance. ScoreCAM image perturbations are created by masking the original input with each feature map \cite{Wang20}. The importance weights are then based on the forward output score of the network for the perturbed image. Conversely, AblationCAM zeroes out image regions that are activated in feature maps \cite{Desai20}. The reduced performance based on the perturbed image is then used as the \(\alpha_k^c\) importance weights for AblationCAM.

\textbf{Non-Discriminative Approaches:} In addition to gradient saturation, the calculation of gradients can be time-consuming and dependent on correct image classification. EigenCAM eliminates these problems and the need for class discrimination by computing the principal components of the feature maps (this method does not follow the general CAM formulation in Eq. \ref{eqn:cam-form}) \cite{Muhammad20}.

\textbf{Non-CAM Gradient-Based Approaches:} While our study proposes MetaCAM, the combination of any CAM-based approach, it can also be extended into non-CAM feature maps. We demonstrate this with the inclusion of FullGrad, which is a non-discriminative method that uses the summed gradients of all bias terms throughout a CNN \cite{Srinivas19}. We include FullGrad and any other visual explanation methods under the term `CAM' throughout this paper.

\subsection{Quantitative Evaluation}

In order to evaluate performance and accuracy of CAM methods, there have been numerous quantitative metrics proposed. Here, we summarize three of the most commonly used metrics for CAM evaluation.

\textbf{Perturbation Analysis:}
One method of evaluating the quality of CAM visualizations is by perturbing the original image with the outputted activations. Some papers choose to mask the least-activated regions from CAM in the original image \cite{Chattopadhay18,Desai20,Srinivas19,Wang20}, while others choose to perturb the most-activated regions \cite{Fu20,Jiang21, Selvaraju17,Srinivas19}. The perturbated image is then processed by the model, and the resulting classification score is used to determine the drop or increase in confidence specifically attributable to the perturbation.

\textbf{Object Localization and Segmentation:}
Another CAM performance metric is based on object localization/segmentation, demonstrated in \cite{Chattopadhay18,Draelos20,Jiang21,Muhammad20,Selvaraju17,Wang20}. Here, true bounding boxes or segmentation masks are provided to identify objects within an image. The most highly activated pixels in CAM feature maps are converted to a binary mask, and the mask is then compared to the desired bounding box or segmentation mask. This is commonly measured through the intersection over union (IoU).

\textbf{Human Evaluation and Class Discrimination:}
One of the reasons CAM methods have been developed is to instill higher trust in CNN model decisions. Certain studies choose to evaluate this by asking human raters to identify which CAM image (among different CAM methods) is more reliable \cite{Chattopadhay18,Desai20,Draelos20}. Similarly, to evaluate whether CAM methods can correctly identify classes, individuals are asked to select which class is best highlighted in the image \cite{Desai20,Fu20,Selvaraju17}. 

Certain metrics must be used with caution when evaluating CAM performance. A CAM model may use areas of an image outside the desired class to make its prediction and as such, there exists concern as to whether object localization and human evaluation/class discrimination metrics are truly evaluating the accuracy of CAM methods. Draelos and Carin highlight this using an example that water in an image may help provide evidence for the classification of a boat \cite{Draelos20}. Thus, IoU and class discrimination should not be used to identify if CAM methods accurately identify salient regions of an image used for model prediction.

\section{Proposed Method}
In this section, we describe MetaCAM’s formulation; a conceptual overview of our proposed method is depicted in Figure \ref{fig:conceptual-overview}. First, we define the problem and outline several methods of averaging individual CAM methods to develop MetaCAM. Next, we describe how MetaCAM can formulated based on the consensus of CAM methods. Briefly, the top \textit{k}\% most highly activated pixels in agreement across all chosen CAMs are used to create MetaCAM. We also describe how this thresholding technique can be used to refine individual CAMs. Lastly, we define how CAMs are evaluated using the Remove and Debias (ROAD) method.

\subsection{Problem Formulation}
Consider a number of feature map visualizations $n$ generated by different CAM methods. The objective is to determine a method of combining these visualizations to improve upon individual CAM performance. MetaCAM should create an accurate visualization of salient regions used in prediction regardless of selected image, target class, or model.

\subsection{Averaging-Based MetaCAM}
An obvious initial formulation of MetaCAM is to simply take an average of all individual CAM methods. Prior to combining CAM visualizations, all maps are normalized between 0 and 1. CAMs that produce an invalid output for a given image/model are removed prior to calculation. 

\begin{equation}
L^c_{MetaCAM(A)} = \frac{\sum\limits_{n}(L^c_{CAM(A)_n})}{n}
\end{equation}

where \textit{n} is the number of individuals CAMs. If all CAMs included in the formulation provide highly accurate visualizations, averaging across CAMs should refine the activated regions and improve performance. However, if any of the included CAMs activate ‘incorrect’ regions of the image, equally weighting all CAMs will reduce the performance of MetaCAM and yield lower performance than other top-performing CAMs. To diminish the inclusion of poor-performing CAMs, a weighted average can be used:

\begin{equation}
L^c_{MetaCAM(A)} = \frac{\sum\limits_{n}(w_n \cdot L^c_{CAM(A)_n})}{\sum\limits_{n}w_n}
\end{equation}

The weights of each CAM can be determined by any quantitative measure of CAM performance. In this study, we choose the ROAD value (section \ref{sec:metacameval}) to numerically quantify and compare CAMs. In addition to using the ROAD values as weights themselves, transformations can be applied to augment the difference in performance between CAMs. We experiment with min-max, softmax (with initial amplification of ROAD weights to a minimum of $10^1$), and exponential normalization of weights. 

While a weighted average of CAMs may improve MetaCAM performance over equal-weighting, poor-performing CAMs are not entirely removed from the overall formulation of MetaCAM and may still negatively affect performance. For this reason, we opt for a consensus-based MetaCAM formulation.

\subsection{Adaptive Thresholding-Based MetaCAM}

To reduce the impact of poor-performing CAMs, we leverage a consensus-based formulation of MetaCAM. Here, rather than averaging CAMs, MetaCAM is generated using the top-$k$\% of pixels in agreement across all CAM methods. Given the consensus of pixel activations across all methods, regions from individual CAMs that are incorrectly activated will be excluded from the final MetaCAM formulation. The activation maps from all CAM methods are summed, and a threshold is applied where the top-$k$\% of summed activations are used as MetaCAM’s activations, with all other activations set to zero:


$ L^c_{MetaCAM(A_{i,j})} = \left\{ 
  \begin{array}{ c l }
    \sum\limits_{n}(L^c_{CAM(A_{i,j})_n}) & \quad \textrm{if }  \geq t \\
    0                 & \quad \textrm{otherwise}
  \end{array}
\right.$

where \textit{t} is the chosen threshold and \textit{i,j} represent pixel locations. The best-performing threshold for MetaCAM is dependent on a given image, target class, and model. We therefore implement adaptive thresholding, which computes the ROAD performance of MetaCAM at different thresholds and returns the highest-performing resultant activation map. For a fair comparison with individual CAM methods, we extend this adaptive thresholding approach for each component CAM method. Thus, ROAD percentiles are calculated using the top-$k$\% of highly activated pixels in each activation map. 

\subsection{MetaCAM Evaluation}\label{sec:metacameval}
To quantify all CAM methods considered in this study we implement a pixel perturbation method, Remove and Debias (ROAD) \cite{Rong22}. ROAD addresses issues of data leakage found in other pixel perturbation methods and improves computational efficiency \cite{Rong22}. To determine the performance of a CAM visualization, ROAD uses noisy linear imputations to perturb either the most or least activated image pixels. The imputations are applied to individual pixels based on neighbouring pixel values, creating a blurred perturbation of the original image as opposed to masking the pixels entirely. The perturbed image is then evaluated by the network to determine the increase or decrease in prediction confidence. Prediction confidence varies depending on the percent of pixels perturbed; for this reason, we choose to evaluate ROAD at 20\%, 40\%, 60\%, and 80\% of perturbation, taking an average across all percentiles for robust evaluation. We also consider  Gildenblat's proposition to combine the confidence drop when the most activated pixels are perturbed, and the confidence increase when the least activated pixels are perturbed \cite{Gildenblat21}. Our final ROAD evaluation score is thus: 


\begin{equation}
ROAD(L^c_{CAM}) = \sum\limits_{p} \frac{(C^p_{L_{LRP}} – C^p_{L_{MRP}})}{2}
\end{equation}

where \textit{p} is the percentile, \textit{C} is the confidence, \textit{LRP} represents least relevant pixels, and \textit{MRP} represents most relevant pixels.

\section{Experiments}
In this section, we outline the experiments used to evaluate MetaCAM. Briefly, we first describe the image dataset, element classes and models, and experimental setup used to evaluate MetaCAM's performance in sections \ref{sec:datamodel} and \ref{sec:exptsetup}. Next, we test different combinations of CAMs (CAM-sets) to determine both the optimal formulation of MetaCAM and top-\textit{k} pixel threshold, evaluated using a novel method denoted Cumulative Residual Effect (CRE) and normalized ROAD scores in sections \ref{sec:systematic}, \ref{sec:ideal-metacam}, \ref{sec:cre}, \ref{sec:comparison} \ref{sec:bad-cams}. Lastly, we compare MetaCAM to individual CAMs both quantitatively and qualitatively in sections \ref{sec:metawin}, \ref{sec:topk}.

\subsection{Datasets \& Pre-Trained Models} \label{sec:datamodel}
To compare MetaCAM performance to other individual CAM methods we selected various images from the ImageNet ILSVRC 2012 validation dataset \cite{deng2009imagenet} in addition to sample images commonly considered as benchmarks for CAM method comparison. Each of the 2D RGB natural images with ground truth segmentation maps has a corresponding numerical class ID. Images are preprocessed by reshaping to an initial size of $256 \times 256$ and center cropping to $224 \times 224$, followed by normalization. For simplicity, a given image (\textit{e.g.} $x$), class (\textit{e.g.} $c$), and model (\textit{e.g.}$f(\cdot)$) are hereafter denoted as an $(x,c,f(\cdot))$ 3-tuple. We use ResNet152 \cite{He15} and DenseNet161 \cite{Huang16} models (unless otherwise specified) pre-trained on the ImageNet-1K dataset from PyTorch \cite{TorchVision16}, comprised of 1,000 possible classes. 

\begin{figure}
    \centering
    \includegraphics[width=0.45\textwidth]{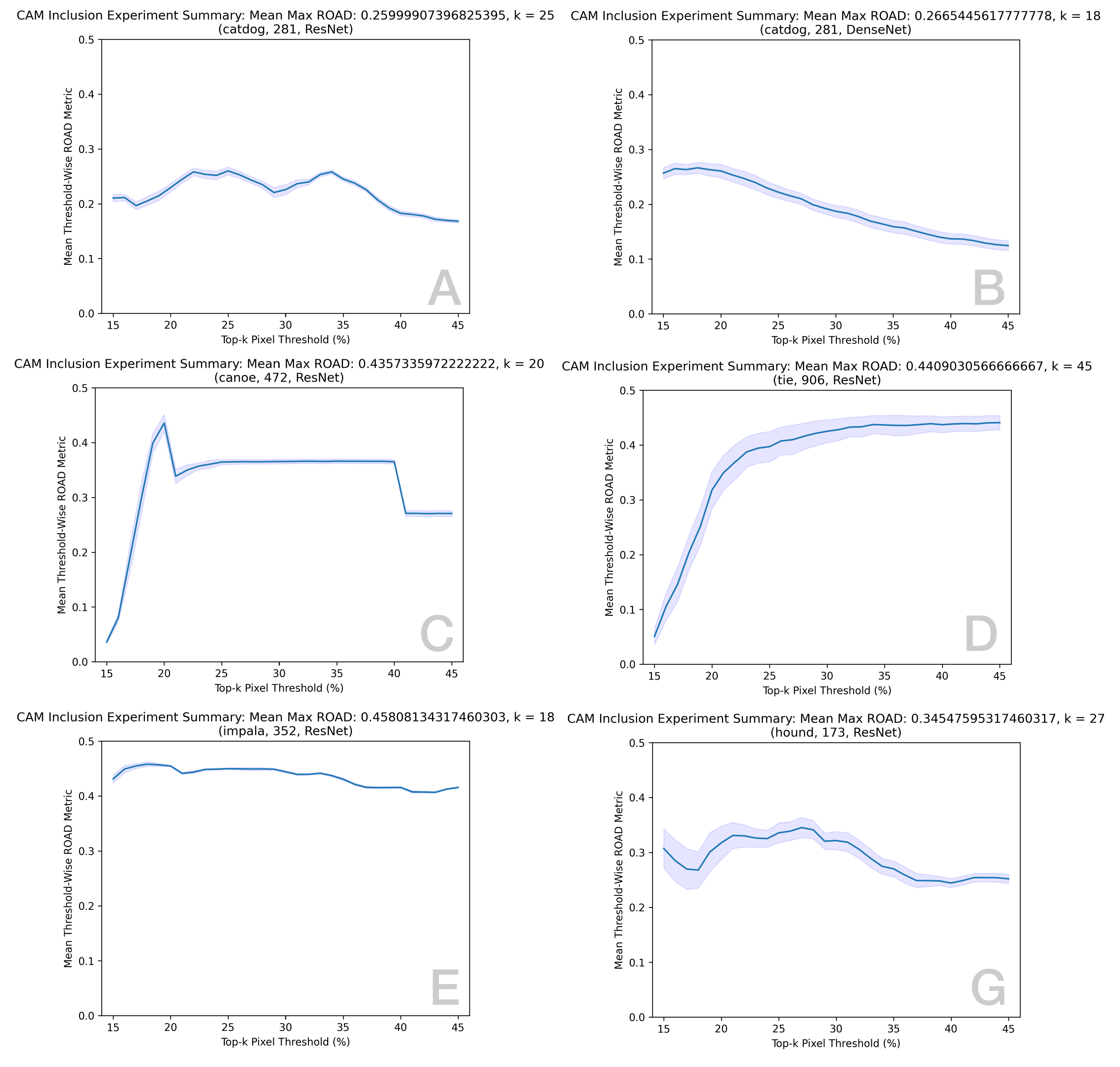}
    \caption{CAM inclusion study curves summarising the adaptive thresholding performance across $m=64$ binary experiments. The mean performance is banded by threshold-wise 95\% confidence intervals.}
    \label{fig:summary-curves}
\end{figure}

\begin{figure}
    \centering
    \includegraphics[width=0.45\textwidth]{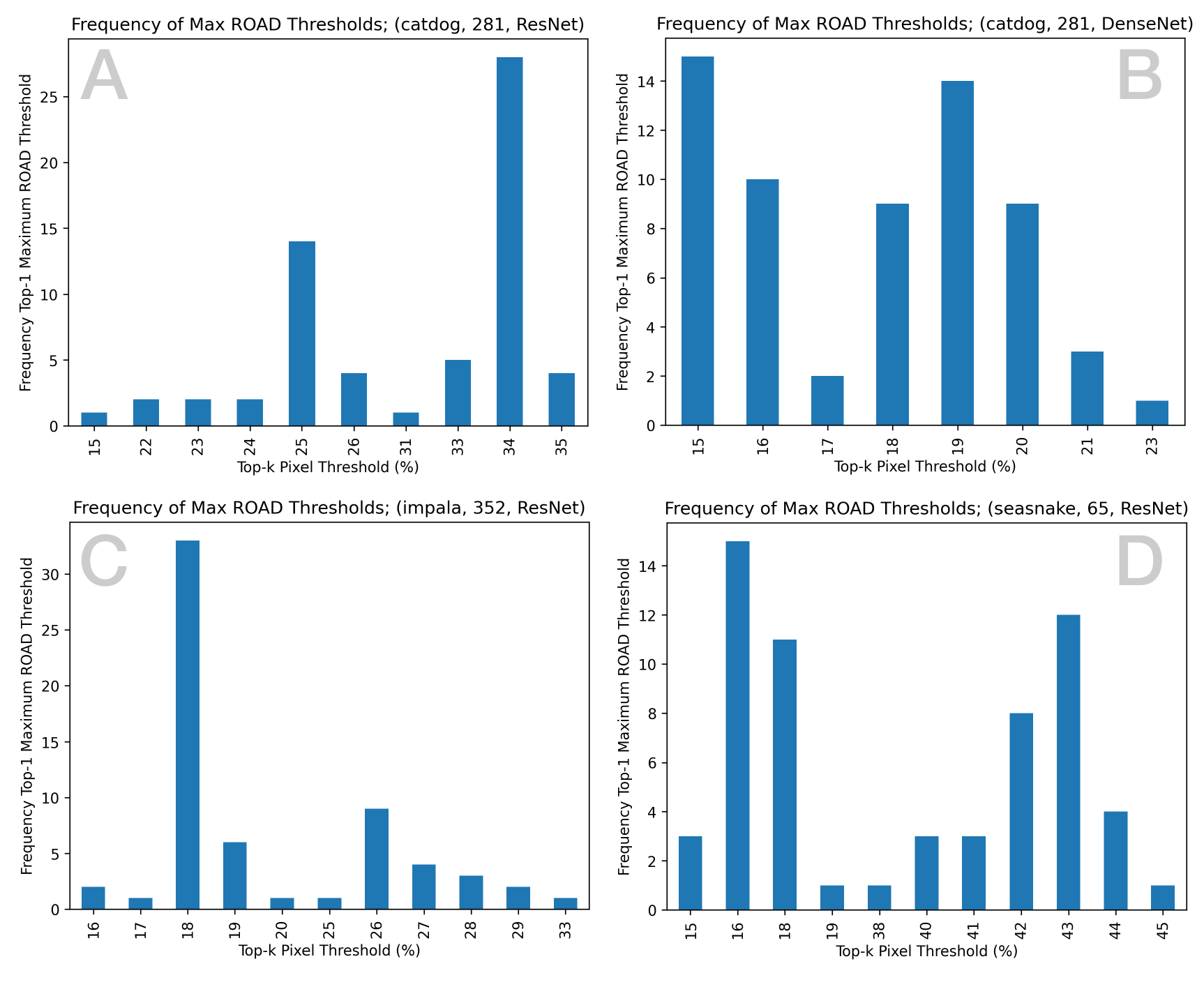}
    \caption{Frequency for which a given top-$k$\% pixel threshold produced the maximum ROAD score across $m=64$ binary experiments.}
    \label{fig:freq-thres-top1}
\end{figure}

\subsection{Experimental Setup} \label{sec:exptsetup}

To experimentally determine what ideal combination of individual CAMs produces the best ROAD score for a given $(x,c,f(\cdot))$ we might trivially consider all available CAMs and/or vision explanation methods. While an initial starting point, such an approach comes at a maximal computational expense and does not provably guarantee optimal results. Rather, as much ensemble-based research is conducted, the optimal combination of components must be determined experimentally. To that end, we leveraged a systematic approach for determining what CAMs to include/exclude in a particular MetaCAM application to a given $(x,c,f(\cdot))$. 

\subsection{Systematically Determining the Ideal CAM-set for MetaCAM} \label{sec:systematic}

The systematic and unbiased methodology to determine what CAM combination produces the optimal MetaCAM result can be determined experimentally. To that end, we leverage high-performance-computing (HPC) infrastructure to uniquely explore the hyper-parameterized inclusion-/exclusion-space of all available CAM methods.

For the $n=11$ individual CAM methods available under consideration, we opted for six unique groupings of methodologies to reduce the computational expense of our systematic experiments ($m=2^{11}$ unique experiments is computationally taxing). Ultimately, we organized all CAM methods into six groupings based on methodological similarity, as reflected in the introduction; certain CAMs were excluded from these large-scale experiments given their individual computational expense. Consequently, we grouped all CAM methods into $n=6; m=2^{n}$ individual experiments that represent the inclusion/exclusion of specific groups: 
\begin{enumerate}[nolistsep]
    \item[] A : [HiResCAM, GradCAMElementwise],
    \item[] B : [GradCAM, GradCAM++],
    \item[] C : [XGradCAM],
    \item[] D : [AblationCAM, ScoreCAM],
    \item[] E : [LayerCAM],
    \item[] F : [FullGrad]
\end{enumerate}

Thus, all systematically determined experiments are summarized through a CAM-specific binary number where $100000$ position-wise represents only the CAMs of group A (1 indicates inclusion only for the binary position for group A). Similarly, 111111 represents the inclusion of all CAMs in groups ABCDEFG. Our work quantifies all ROAD performance of the $m=64$ individual experiments for a given $(x,c,f(\cdot))$. The maximum achievable ROAD score is determined across all $m=64$ experiments and a relative ranking of CAM inclusion/exclusions is further explored in subsequent sections. Individual experiments were allocated $4 \times$ cores, 64GB of RAM, and either an NVIDIA P100 Pascal or V100 Volta GPU through the Digital Research Alliance HPC infrastructure.

\begin{figure}
    \centering
    \includegraphics[width=0.45\textwidth]{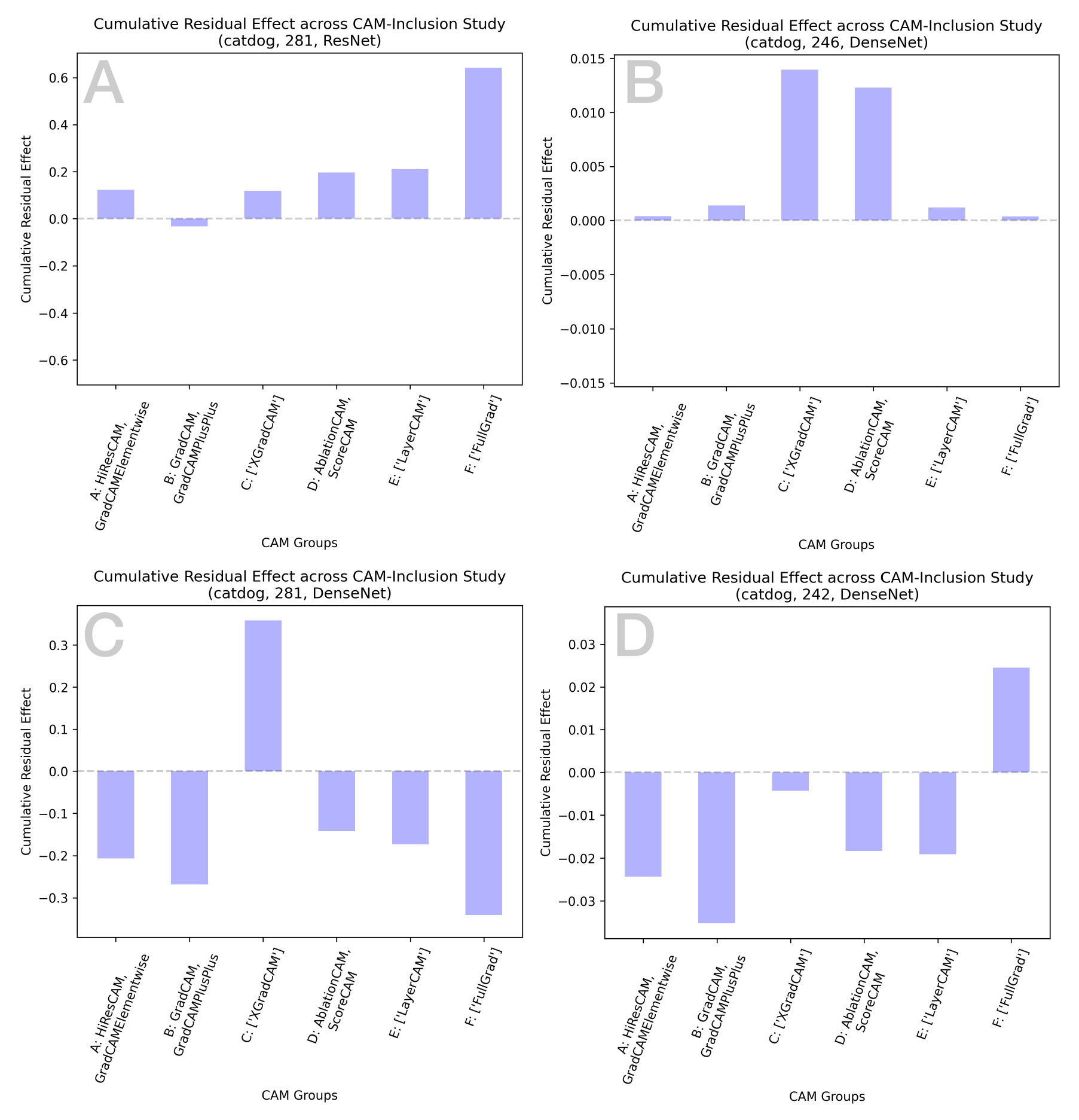}
    \caption{Cumulative Residual Effect summarizing group-wise impact on MetaCAM performance}
    \label{fig:cre}
\end{figure}

\begin{figure}
    \centering
    \includegraphics[width=0.45\textwidth]{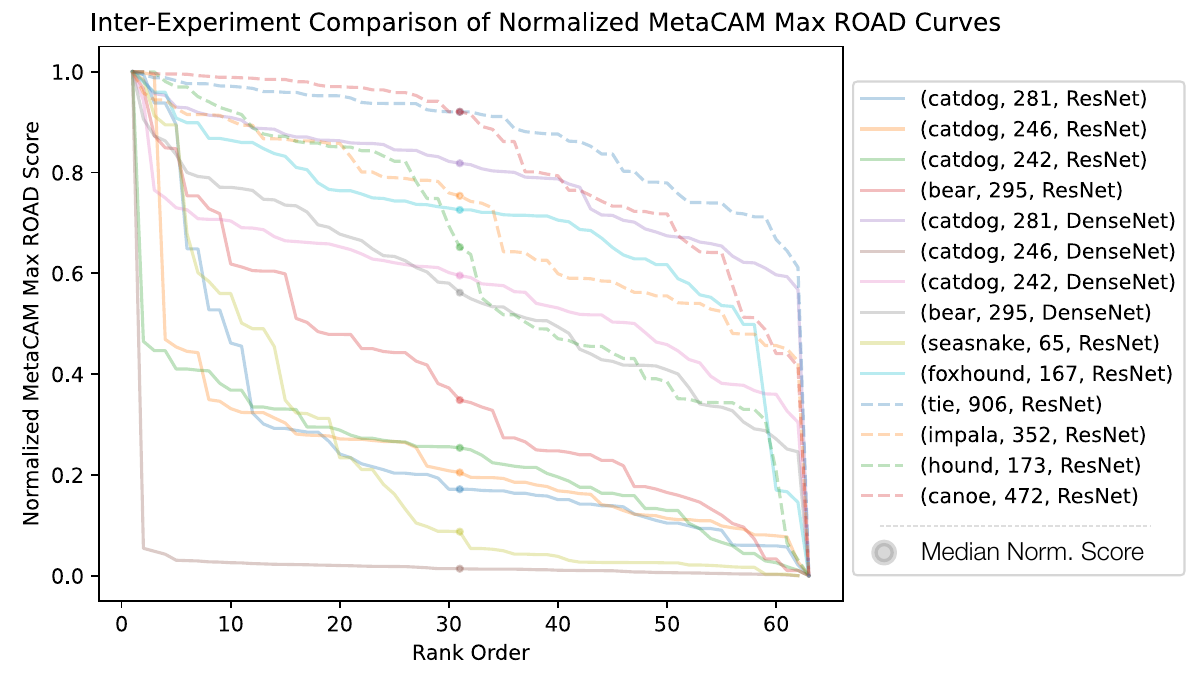}
    \caption{Comparison of normalized inter-experiment maximum ROAD scores.}
    \label{fig:norm-inter-expt}
\end{figure}

\begin{figure*}
    \centering
    \includegraphics[width=\textwidth]{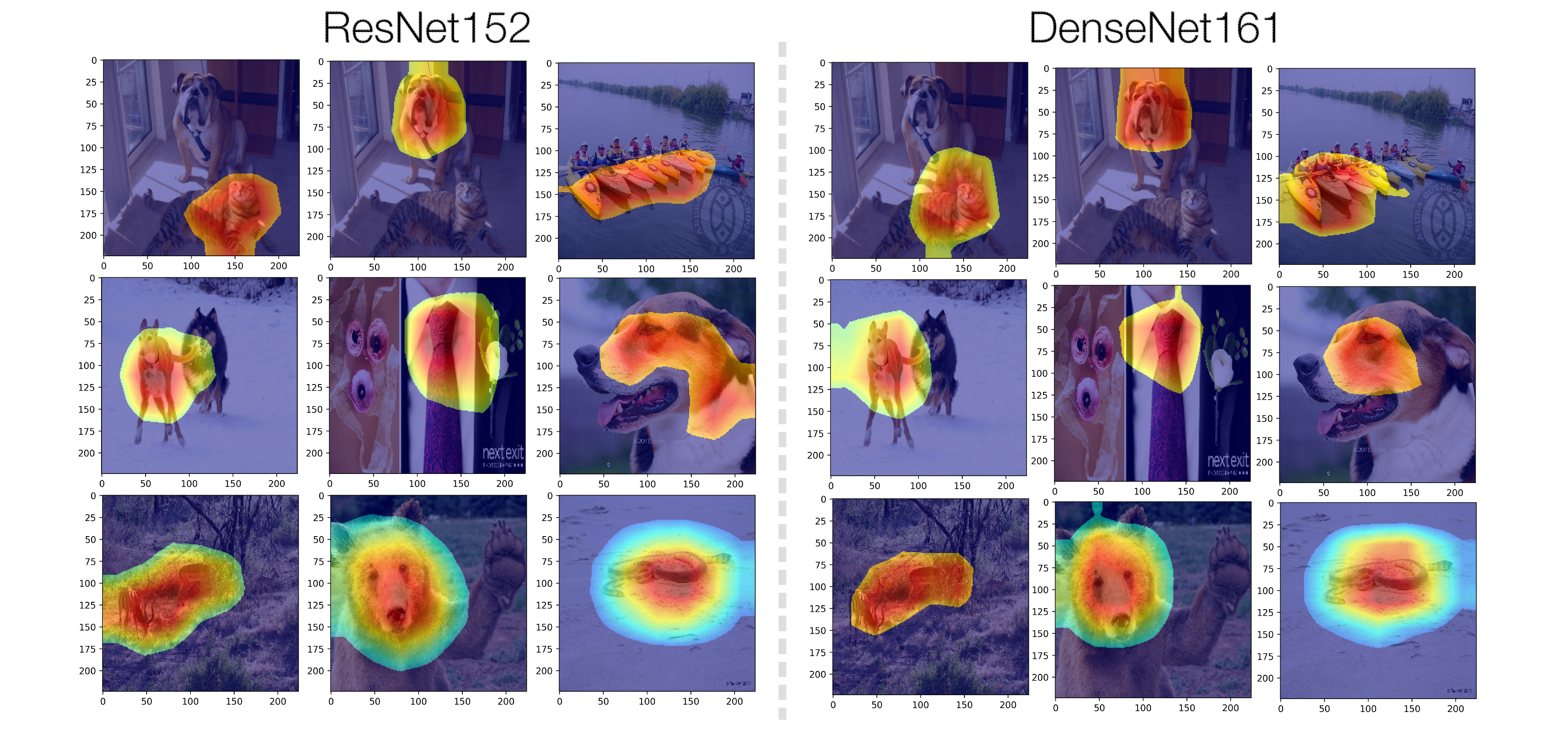}
    \caption{Visualization of the MetaCAMs from the best performing threshold across various images, classes, \& model architectures.}
    \label{fig:best-metacam}
\end{figure*}

\begin{figure}
    \centering
    \includegraphics[width=0.5\textwidth]{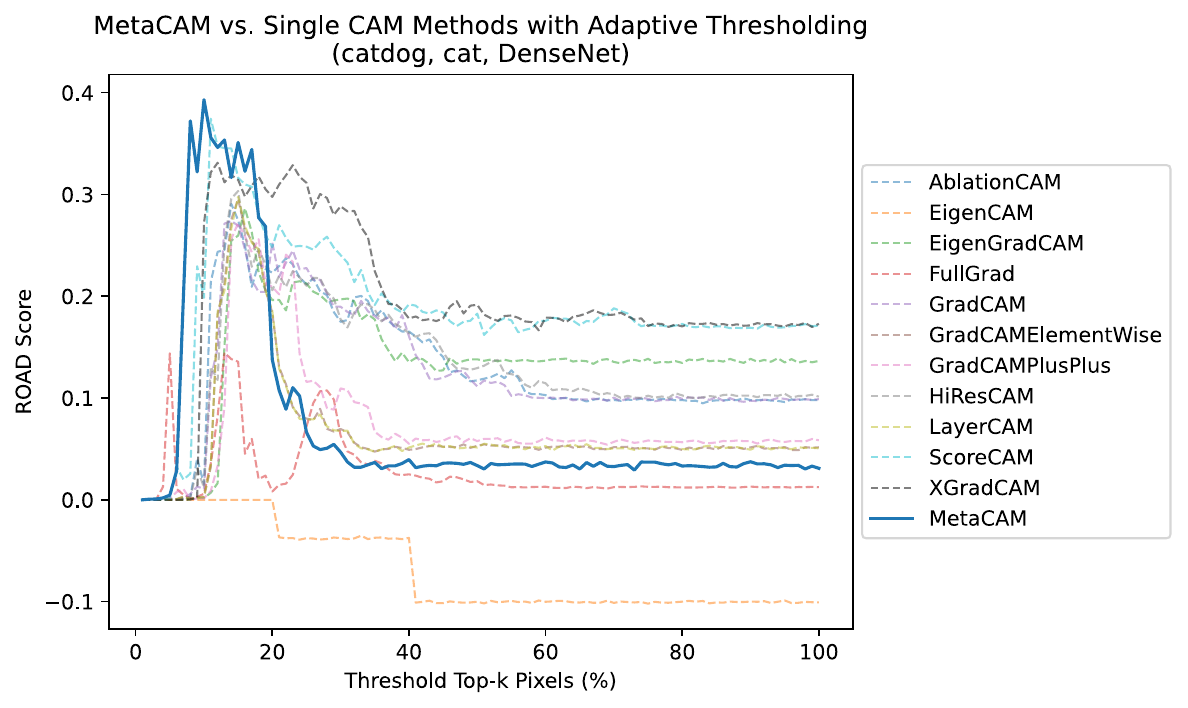}
    \caption{MetaCAM outperforms individual methods even when augmented through adaptive thresholding.}
    \label{fig:metacam-wins}
\end{figure}

\begin{figure*}
    \centering
    \includegraphics[width=\textwidth]{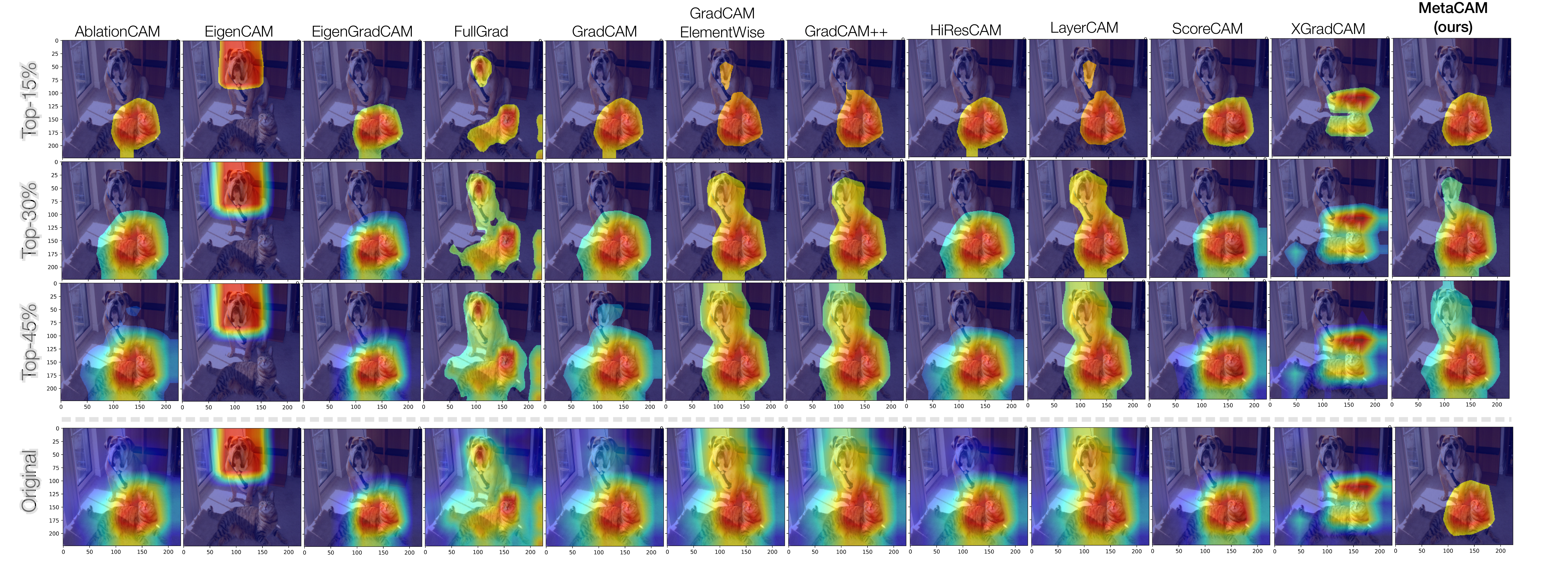}
    \caption{Comparison of individual CAMs using a pre-trained DenseNet161 and the cat class ID (281).}
    \label{fig:compare-cams}
\end{figure*}

\subsection{Experimentally Determining Ideal MetaCAM Top-$k$\% Pixel Threshold} \label{sec:ideal-metacam}

The proposed MetaCAM method leverages top-$k$ pixel adaptive thresholding that can be experimentally determined across the $m=64$ individual experiments. For a given range of values $k$ (typically $k \in [15,45]$, determined experimentally) the mean ROAD score is depicted for various $(x,c,f(\cdot))$ in Figure \ref{fig:summary-curves}. We note that there does not exist a universally applicable threshold range and that in some cases (\textit{e.g.} E) there does not exist a definitive ideal threshold whereas for others (\textit{e.g.} A,B,C,D,G) there are peak threshold or ranges providing dramatic increases in ROAD score.

To further investigate the influence on MetaCAM's top-$k$ thresholding, Figure \ref{fig:freq-thres-top1} depicts a sample of four experiments and the frequency at which a given value $k$ produced the top-1 maximum ROAD score. Interestingly, certain $(x,c,f(\cdot))$ produce a definitive $k$ (\textit{e.g.} A,C) while others suggest a general range for $k$ (\textit{e.g.} B,D). Intuitively, the complexity of the contents of a given image and the classes represented therein are strong determinants of the ideal $k$; images with a clear distinction of class-specific pixels and the remainder of the image do not overly benefit from an adaptive threshold (Figure \ref{fig:summary-curves}E) while all others suggest tuning $k$. As a rule of thumb, a modest amount of top-$k$ thresholding appears to consistently improve ROAD score, often by a considerable margin. 

\subsection{Cumulative Residual Effect} \label{sec:cre}
The $m=64$ binary CAM-group inclusion/exclusion experiments for a given $(x,c,f(\cdot))$ produce a unique relative ranking of experiment-specific binary numbers according to the MetaCAM ROAD score. To quantify the influence a given CAM group had on the resultant MetaCAM ROAD score we propose the Cumulative Residual Effect method. CRE determines the relative positive/negative effect of each CAM group by taking the residual of the individual MetaCAM score with the median value of all scores for the $m=64$ experiments and sum this residual (either a positive or negative value) with each contributing CAM groups within that experiment. This produces a group-wise summary representing the relative impact of including/excluding that CAM group; all CAM groups are included in exactly 32 experiments and excluded in exactly 32 experiments.

Since this method leverages residuals, it has an additional useful property of being aggregated with other experiments; that is, inter-experiment CREs are obtained by summing intra-experiment group-wise values. In Figure \ref{fig:cre} we illustrate four example CRE plots that may be roughly interpreted in a way analogous to SHAP force plots \cite{lundberg2017unified}. The inclusion of certain CAM groups will positively improve MetaCAM ROAD score (\textit{e.g.} A,B) relative to the median of all experiments, and in other cases, their inclusion negatively influences overall ROAD scores (\textit{e.g.} C,D; it is critical to note that this quantification is relative to the experiment-wide median ROAD score).

\subsection{Comparison of MetaCAMs across Images, Classes, \& Models} \label{sec:comparison}

To fully depict the MetaCAM output across varying $(x,c,f(\cdot))$, we visualize the best-performing MetaCAMs by ROAD score in Figure \ref{fig:norm-inter-expt}. We note that our consensus-based approach reliably detects the target class in all cases with dramatic improvements in ROAD score over individual CAMs (section \ref{sec:metawin}).

\subsection{``Bad'' CAMs \& RandomCAM can Improve Performance} \label{sec:bad-cams}
Surprisingly, the inclusion of individually poor-performing CAMs (such as EigenCAM which often resulted in low or negative ROAD scores) and random noise (\textit{e.g.} RandomCAM, an activation map generated from a random uniform distribution between -1 and 1 \cite{Gildenblat21}) resulted in improved MetaCAM performance. Our initial top-performing MetaCAM combination using DenseNet161 to predict the cat in the catdog image achieved ROAD performance of 0.295. However, the inclusion of EigenCAM and RandomCAM improved ROAD performance to 0.393 and 0.314, respectively. In addition, the highest-performing threshold decreased from 19 to 10 with EigenCAM, and 15 with RandomCAM. We posit that the inclusion of poor-performing CAMs or random noise forces MetaCAM to further refine its output to only those highest-consensus pixels at lower top-$k$ threshold values. Giving random or incorrect regions higher activations results in a refinement of the specific area used in model predictions and contributes to the exclusion other classes within the same image. Interestingly, future work might consider incorporating various ``bad"/random-based CAMs to experimentally determine their impact on overall MetaCAM performance.

\subsection{MetaCAM Outperforms Individual Methods} \label{sec:metawin}

To compare the performance of MetaCAM against other visual explanation methods, we applied adaptive thresholds to each individual visualization, displayed in Figure \ref{fig:metacam-wins}. MetaCAM outperforms all individual CAM methods, shown by the largest peak at $k=10$.  Most CAMs have a peak performance using a threshold between the top 10\%-30\% of most highly activated pixels. This indicates that adaptive thresholding is able to improve performance of all visual explanation methods, by selecting the top relevant pixels for a given $(x,c,f(\cdot))$. In addition, Figure \ref{fig:metacam-wins} also shows the original performance of CAMs without adaptive thresholding ($k=100$). Here, ROAD performance ranges between -0.101-0.172, further demonstrating that MetaCAM (ROAD=0.393) has a dramatic increase in performance compared to original CAMs.

\subsection{Comparison of CAMs across different \textit{k}\% Pixel Thresholds} \label{sec:topk}

Figure \ref{fig:compare-cams} displays our adaptive thresholding technique applied to all 11 CAMs explored in this study, in addition to MetaCAM. CAMs are shown using thresholds of the top 15\%, 30\%, and 45\% activated pixels, as well as the original CAM visualization. These qualitative results present the usefulness of thresholding; reducing the number of activated pixels focuses on the most salient regions of each image. Many of the original CAM visualizations activate large regions of the image, including both the cat and dog despite only using the cat class ID (281) as the target. Adaptive thresholding is able to refine the activations of all CAMs to focus on the desired target class. It is important to note that thresholding is not performing object localization; it is focusing on the regions of an image used by a model for prediction, as measured by ROAD.


Interestingly, EigenCAM incorrectly highlights the dog in the image, instead of the desired cat class. This explains the negative ROAD value for EigenCAM in Figure \ref{fig:metacam-wins}. EigenCAM is a non-discriminative CAM method that uses principal components to create activation maps. However, when there are multiple classes within the same image, the order of prinicipal components must be specified (\textit{e.g.,} first principal components vs. second principal components). EigenCAM performs well on images with a ``single-subject", but otherwise requires a user to determine the number and rank of various components within an image to perform successfully. This requires a level of hand-engineering and data leakage. Despite the potential of including `incorrect' classes in an image, EigenCAM may still be beneficial to the performance of MetaCAM as described previously.



\section{Conclusion}
In this study, we propose MetaCAM, a consensus-based combination of any number of existing CAM formulations using the top \textit{k}\% of pixels in agreement across all methods. Our experiments demonstrate that MetaCAM is able to outperform existing CAM methods, both with and without adaptive thresholding. We expect MetaCAM to be of particular use in high-criticality fields.

\section*{Acknowledgment}
The authors would like to acknowledge Dr. Katherine Muldoon for her support of this work. The authors also acknowledge that this study took place on unceded Algonquin Anishinabe territory.

{\small
\bibliographystyle{ieee_fullname}
\bibliography{references}
}

\end{document}